\documentclass{article}

\usepackage{arxiv}

\usepackage[utf8]{inputenc} % allow utf-8 input
\usepackage[T1]{fontenc}    % use 8-bit T1 fonts
\usepackage{hyperref}       % hyperlinks
\usepackage{url}            % simple URL typesetting
\usepackage{booktabs}       % professional-quality tables
\usepackage{amsfonts}       % blackboard math symbols
\usepackage{nicefrac}       % compact symbols for 1/2, etc.
\usepackage{microtype}      % microtypography
\usepackage{lipsum}		% Can be removed after putting your text content
\usepackage{graphicx}
\usepackage{natbib}
\usepackage{doi}
\usepackage{bbding}
\usepackage{amsmath}
\usepackage{hyperref}
\usepackage{url}
\usepackage{multirow}
\usepackage[table,xcdraw]{xcolor}
\usepackage{graphicx}
\usepackage{times}
\usepackage{latexsym}
\usepackage{geometry}

\newtheorem{Def}{Definition}

\title{\textbf{TSO}: Self-Training with Scaled Preference Optimization}

\author{Kaihui Chen, Hao Yi, Qingyang Li\thanks{Corresponding author.}, Tianyu Qi, Yulan Hu, Fuzheng Zhang\\
	Kuaishou Technology, Beijing, China\\
 \texttt{chenkaihui,yihao,liqingyang,qitianyu03,huyulan,zhangfuzheng@kuaishou.com}\\
    \AND
     Yong Liu\\
	Renmin University of China, Gaoling School of Artificial Intelligence, Beijing\\
\texttt{liuyonggsai@ruc.edu.cn}\\
}
          
\renewcommand{\shorttitle}{\textbf{TSO}:Self-Training with Scaled Preference Optimization}

\hypersetup{
pdftitle={A template for the arxiv style},
pdfsubject={q-bio.NC, q-bio.QM},
pdfauthor={David S.~Hippocampus, Elias D.~Striatum},
pdfkeywords={First keyword, Second keyword, More},
}

    \makeatletter
\def\@fnsymbol#1{\ensuremath{\ifcase#1\or \dagger\or \ddagger\or
   \mathsection\or \mathparagraph\or \|\or **\or \dagger\dagger
   \or \ddagger\ddagger \else\@ctrerr\fi}}
    \makeatother

\begin{document}

\maketitle
\begin{abstract}
Enhancing the conformity of large language models (LLMs) to human preferences remains an ongoing research challenge. Recently, offline approaches such as Direct Preference Optimization (DPO) have gained prominence as attractive options due to offering effective improvement in simple, efficient, and stable without interactions with reward models. However, these offline preference optimization methods highly rely on the quality of pairwise preference samples. Meanwhile, numerous iterative methods require additional training of reward models to select positive and negative samples from the model's own generated responses for preference learning. 
%It is still challenging to continuously construct high-quality positive and negative preference instances as the ability of LLM improves without an explicit reward model.
Furthermore, as LLMs' capabilities advance, it is quite challenging to continuously construct high-quality positive and negative preference instances from the model's outputs due to the lack of diversity. 
%Self-Rewarding employs the language model,  which is utilized through LLM-as-a-Judge prompting, to generate its own rewards throughout the training process. 
%High-quality data requirements for preference alignment  include \textit{diversity}, \textit{validity} and \textit{adaptability}. 
To tackle these challenges, we propose TSO, or \textit{Self-Training with Scaled Preference Optimization}, a framework for preference optimization that conducts self-training preference learning without training an additional reward model. TSO enhances the diversity of responses by constructing a model matrix and incorporating human preference responses. Furthermore, TSO introduces corrections for model preference errors through human and AI feedback. Finally, TSO adopts iterative and dual clip reward strategies to update the reference model and its responses, adaptively adjusting preference data and balancing the optimization process. Experimental results demonstrate that TSO outperforms existing mainstream methods on various alignment evaluation benchmarks, providing significant insight into preference data construction and model training strategies in the alignment domain.
\end{abstract}

\section{Introduction}

 Reinforcement Learning from Human Feedback (RLHF) has emerged as an effective method to fine-tune Large Language Models (LLMs) to align better with human users' expectations~\cite{schulman2017proximal,rafailov2024direct,achiam2023gpt}. It utilizes algorithms like Proximal Policy Optimization (PPO, ~\citet{schulman2017proximal}) and Direct Preference Optimization (DPO, ~\citet{rafailov2024direct}). While PPO is known for its relatively good sample efficiency, it is challenging to train online and demands extensive tuning of hyperparameters. On the other hand, DPO is a lightweight and offline algorithm that directly optimizes policies, offering greater flexibility and easier implementation compared to PPO.

However, offline preference optimization methods highly rely on the quality of pairwise preference samples. It is still challenging to continuously construct high-quality positive and negative preference instances as the ability of LLM improves without an explicit reward model. To address these issues, we propose that high-quality preference alignment data should satisfy diversity, validity, and adaptability. Diversity entails that prompts cover a wide range of topics, languages, and tasks, and that responses are sampled from a plenty of various distribution, particularly regarding negative responses. Humanity requires that the preferences in responses undergo correctness verification by humans or AI to mitigate the noise generated by out-of-distribution (OOD) instruction data, affecting the model's alignment effectiveness. Adaptability implies that as the target model updates, its responses are supposed to be promptly updated to eliminate misleading signals from the old model. Previous work, such as PRO ~\cite{song2024preference}, focuses only on the data's diversity and validity, neglecting adaptability because the model's  responses are not updated promptly during the optimization process; Self-Reward ~\cite{yuan2024self} has adaptability but lacks validity and diversity.Relying solely on controlling temperature to change the diversity of the distribution is limited and model evaluation is not corrected by human feedback, which could evaluate OOD response wrongly. For a more detailed comparison of the preference alignment data properties across different methods, refer to the Table ~\ref{tab:preference-property}.

\begin{figure*}[htb]
    \centering
    \includegraphics[width=0.9\linewidth]{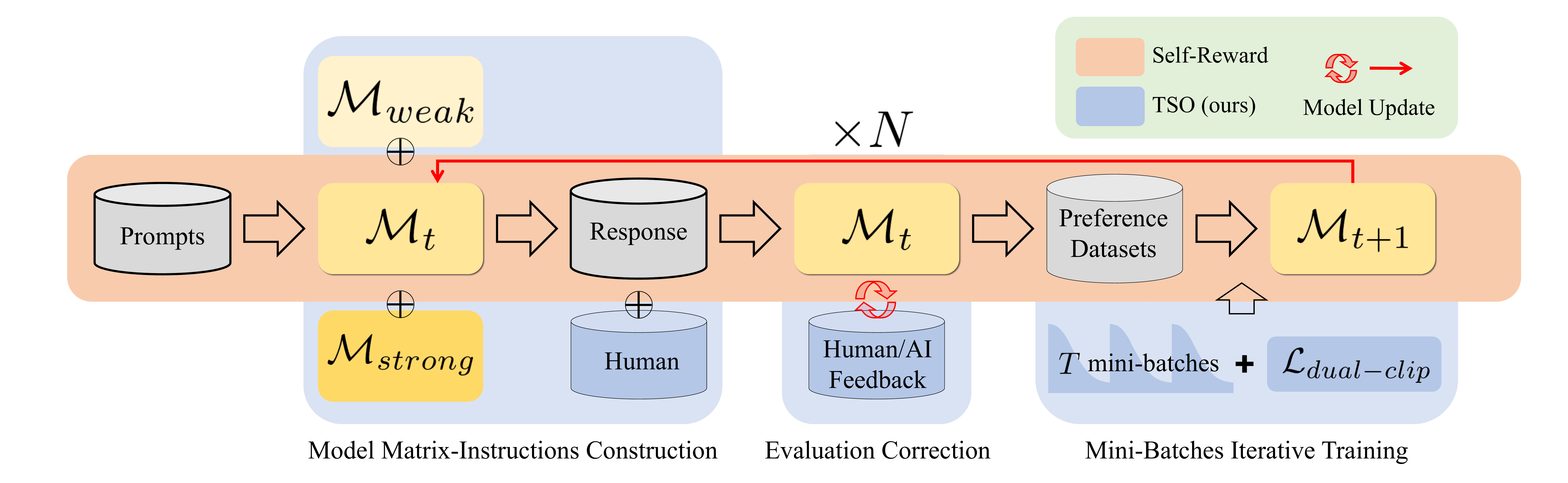}
    \caption{TSO first samples responses from the model matrix, ensuring the diversity of the positive and negative response datasets. Then, it uses feedback from humans or AI to correct validity bias. Finally, it employs the Mini-Batch Iterative DPO and Dual Clip Reward Loss strategies for DPO training. The above steps are repeated $N$ times.}
    \label{fig:TSO}
\end{figure*}

 To simultaneously balance the diversity, validity, and adaptability of preference alignment data, we propose a multi-stage self-training framework, called Self-\textbf{T}raining with \textbf{S}caled Preference \textbf{O}ptimization (\textbf{TSO}), which includes \textit{model matrix instructions construction}, \textit{evaluation correction}, and \textit{mini-batches iterative DPO training}, as shown in Figure \ref{fig:TSO}. The \textit{model matrix instructions construction} stage initially constructs instruction response samples through the model matrix, leveraging data diversity, especially the negative responses, to enhance the model's generalization capability and efficiency. During the \textit{evaluation correction} stage, human and AI feedback are applied to continuously correct validity bias in the evaluation process and improve the handling of out-of-distribution (OOD) samples, thus ensuring data validity. Finally, in the \textit{mini-batches iterative DPO training} stage, by partitioning the dataset and switching the reference model, we improve the efficiency of data utilization and enhance the alignment performance. Additionally, we revised the original DPO loss function and introduced the dual clip reward loss, which effectively mitigates the imbalance between positive and negative samples during the optimization process, thus ensuring the adaptability of both data and the optimization process. The contributions of this work are summarized as follows:
\begin{itemize}
    \item We propose a self-training framework, constructing preference data by considering diversity, validity, and adaptability during the iterative learning process.
    \item We introduced model update strategies that involve mini-batches iterative DPO and dual clip reward loss, which improved the efficiency of data utilization, balancing the optimization between positive and negative responses.
    %\item The robustness of this training paradigm is validated on a rich set of alignment evaluation sets, and compared to existing preference alignment algorithms, the alignment performance and effectiveness have been greatly enhanced.
    \item We explored the relationship between the model alignment effect and the distribution of positive and negative preference instances, providing practical insights into pairwise preference data construction.
\end{itemize}

\begin{table}[t]
\centering
\begin{tabular}{c|ccc}

\hline
Method/Property & D & V & A \\ \hline
DPO ~\cite{rafailov2024direct}            & \Checkmark         & \XSolidBrush            & \XSolidBrush            \\
IPO ~\cite{azar2024general}            & \Checkmark         & \XSolidBrush            & \XSolidBrush            \\
RSO   ~\cite{liu2023statistical}        & \XSolidBrush         & \Checkmark            & \XSolidBrush            \\
ReST ~\cite{gulcehre2023reinforced}           & \XSolidBrush         & \Checkmark            & \Checkmark            \\
RRHF ~\cite{yuan2023rrhf}           & \Checkmark         & \Checkmark            & \XSolidBrush            \\
RPO  ~\cite{song2024preference}           & \Checkmark         & \XSolidBrush            & \XSolidBrush            \\
RAFT ~\cite{dong2023raft}            & \XSolidBrush         & \Checkmark            & \XSolidBrush            \\
Self-Reward  ~\cite{yuan2024self}   & \XSolidBrush         & \XSolidBrush            & \Checkmark            \\
\textbf{TSO(ours)}            & \Checkmark         & \Checkmark            & \Checkmark            \\ \hline
\end{tabular}

\caption{Summary of the property of preference data used in existing preference alignment solutions.``D'',  ``V'', and ``A'' denote diversity, validity, and adaptability, respectively. }
\label{tab:preference-property}
\end{table}

\section{Related Work}
% \textbf{Best of N (BoN)} ~\citet{stiennon2020learning} \& ~\citet{nakano2021webgpt} an potent strategy for aligning human preferences during the decoding phase. BoN functions by repeatedly sampling $N$ times and scoring each sample using a reward model, ultimately selecting the response with the highest score as the system's output. Previous work ~\cite{lambert2023alignment,pan2022effects} have successfully mitigated the KL divergence from the reference model by decreasing the number of samples $N$, which has led to a decline in model performance. ~\citet{jinnai2024regularized} introduced proximity regularization, which has effectively reduced the risk of over-fitting in the reward model.

 \textbf{RLHF $\&$ RLAIF} Although instruction fine-tuning and SFT can somewhat enhance the models' alignment with human preferences, these methods are heavily dependent on the quality of data, which incurs significant time and monetary costs ~\cite{yao2023deepspeed,touvron2023llama}. Alignment techniques, such as RLHF and RLAIF,  
 leverage human or AI feedback to modulate and steer LLMs' behavior, thereby enhancing LLMs' comprehension of human requirements and refining their responses for improved alignment. ~\citet{ouyang2022training} employ a reward model derived from preference data as the reward function in the actor-critic approach, utilizing reinforcement learning techniques like Proximal Policy Optimization (PPO, ~\citet{schulman2017proximal})  to optimize the target policy.  ~\citet{rafailov2024direct}  uses the target policy to typify the optimal reward function and training it directly using preference data. IPO~\cite{azar2024general} robustly curtails over-fitting in DPO by managing the logarithmic ratio rewards' variance. Conditional DPO (cDPO) considers the inherent noise in preference labels and the probability that high-quality samples surpass low-quality samples being less than one. RSO ~\cite{liu2023statistical} applies rejection sampling to achieve a more precise estimation of the optimal strategy.  ~\citet{lee2023rlaif} utilize AI-generated feedback as a substitute for human feedback to broaden and expedite the language model's alignment process with human preferences.

 \textbf{Self-Training} endeavors to utilize the model's responses, augment and enhance training data quality, aiming for continual alignment with targeted human preferences through inherent capabilities. ReST~\cite{gulcehre2023reinforced} implements a bifurcated approach termed ``Grow'' and ``Improve'', conducting iterative updates to the model. RAFT~\cite{dong2023raft} utilizes the Best of N (BoN, ~\citet{stiennon2020learning},  ~\citet{nakano2021webgpt}) strategy, sampling various responses from the target model and scoring them with the reward model to select the best response, and use SFT to train the target model. RRHF~\cite{yuan2023rrhf} utilizes the conditional probabilities of sample responses from different sources to align with human preferences with a ranking loss. Self-Reward~\cite{yuan2024self} fabricates various responses for identical queries via the target model, scores these using the same model, and builds a preference dataset based on the model's scoring results, iteratively conducting DPO training. PRO~\cite{song2024preference} produces diverse responses and then high-scoring responses and all low-scoring responses are combined into a preference dataset.

\section{Self-Training with Scaled Preference Optimization (TSO)}\label{sec:lpo}

 % Our method is similar to the previous work ~\cite{yuan2024self}, requiring the model to possess dual capabilities: response capability and command creation ability. These are used respectively for generating helpful, harmless, honest replies, and for evaluating these responses to create preference datasets, which are then subject to cyclical iteration to continually enhance the model's capabilities.

 % As illustrated in the Figure ~\ref{fig:TSO}, our method primarily optimizes three aspects: \textit{Model Matrix Instructions Construction}, \textit{Evaluation Correction}, and \textit{Mini-Batches Iterative DPO}. These are aimed at enhancing the diversity of response candidates, correcting erroneous estimations that occur during the evaluation process, and improving training efficiency.

 % In the \textit{Model Matrix Instructions Construction}, the Model Matrix expands the diversity of the response dataset by obtaining model responses that are either stronger or weaker than the target model, through historical versions and different model sizes,  improving model efficiency and effectiveness. In the \textit{Evaluation Correction}, by collecting new responses generated during the model iteration process and acquiring human and AI feedback, the corrective data is served to update and refine the model, continually enhancing its evaluation capabilities. In the \textit{Mini-Batches Iterative DPO}, we introduce mini-batches update strategies and Dual Clip Reward Loss, and train the model iteratively.

In this section, we first outline the process of creating preference pairs using the model matrix, which involves cross-version response augmentation and cross-scale response augmentation. Next, we introduce human and AI feedback to correct validity bias in base model. Finally, we discuss our training strategy, the mini-batches iterative DPO, and introduce dual clip reward loss to balance the optimization process for both positive and negative responses.

\subsection{Model Matrix Instructions Construction}\label{sec:lpo-mmic}

As indicated by previous work ~\cite{kaplan2020scaling}, there is a positive correlation between model size and capability. Larger models exhibit better performance, and models of new versions generally outperform older models during the iterative process. Consequently, we start from these two dimensions, integrating the model matrix to further increase the quality of positive instruction responses and the diversity of negative instruction responses.

At first, we introduce the definition of model matrix $\mathfrak{M}$. See Definition ~\ref{def:model_matrix}.
\begin{Def}
\label{def:model_matrix}
$\mathfrak{M}: \{\mathcal{M}_{v,s}\}_{v\in \mathcal{V},s\in \mathcal{S}}$ is a model matrix, where $\mathcal{V}$ denotes the model's version set and $\mathcal{S}$ denotes the model's size set. The element in $\mathfrak{M}$ is  $\mathcal{M}_{v,s}:(\mathcal{X},\mathcal{Y})\rightarrow [0,1]^{L}$ denotes the model distribution of version v and size s, where $\mathcal{X}$ is the set of prompts and $\mathcal{Y}$ denotes the set of responses with a maximum length of L.
\end{Def}

Assuming our base model is $\mathcal{M}_{v_b,s_b}$, we define the model set to generate chosen responses $\mathfrak{M}_w$ and the model set to generate rejected responses $\mathfrak{M}_l$. See Equation \ref{equ:m_w} and Equation \ref{equ:m_l}.

\begin{align}
    \mathfrak{M}_w=\{\mathcal{M}_{v_{max},s_{max}}\} \cup \{\mathcal{H}\}
    \label{equ:m_w}
\end{align}
\begin{align}
    \mathfrak{M}_l=\{\mathcal{M}_{v,s} \in \mathfrak{M} | t< t_b,s< s_b\}
    \label{equ:m_l}
\end{align}
where $v_{max}=\max\{\mathcal{V}\}$ and $s_{max}=\max\{\mathcal{S}\}$, $\mathcal{H}:(\mathcal{X},\mathcal{Y})\rightarrow [0,1]^{L}$ denotes the expected human responses distribution.

After identifying candidate sampling model sets for positive and negative responses, we then define the sampling distributions for positive and negative responses. See Equation \ref{equ:M_w} and Equation \ref{equ:M_l}.

\begin{align}
    \mathcal{M}_w = \sum_{\mathcal{M} \in \mathfrak{M}_w} W_{\mathcal{M}}^{w}\mathcal{M}
    \label{equ:M_w}
\end{align}
\begin{align}
    \mathcal{M}_l = \sum_{\mathcal{M} \in \mathfrak{M}_l} W_{\mathcal{M}}^{l}\mathcal{M}
     \label{equ:M_l}
\end{align}
where $W_{\mathcal{M}}^{w}$,$W_{\mathcal{M}}^{l}$ denote weighting coefficients of distributions and $\sum_{\mathcal{M}} W_{\mathcal{M}}^{w}=1$ and $\sum_{\mathcal{M}} W_{\mathcal{M}}^{l}=1$. In simple terms, for a given prompt, we perform a single sampling from numerous candidate responses based on their weights.

Finally, we generate the positive instruction dataset $\mathcal{D}_{I}^w$ and the negative instruction dataset $\mathcal{D}_{I}^l$ as follows:
\begin{align}
    \mathcal{D}_{I}^w=\{(x,y_w) | x \sim \rho, y_w \sim \mathcal{M}_w\}
\end{align}
\begin{align}
    \mathcal{D}_{I}^l=\{(x,y_l) | x \sim \rho, y_l \sim \mathcal{M}_l\}
\end{align}
where $\rho$ denotes the prompts distribution. To be concise in description, all prompts are sampled from $\rho$ in the following Equations.

In a nutshell, by leveraging models of different versions and sizes throughout the iterative process, we construct a model matrix that generates a variety of response outcomes. Here, largest and newest version models combined with human response are used to produce high-quality positive responses, while smaller and older version models generate diversity negative samples.

\subsection{Evaluation Correction}
 % To avoid erroneous evaluations during the model training process, an Evaluation correction phase is incorporated into our model training workflow. In this phase, preference data obtained from model evaluations is continuously collected, and corrections and fine-tuning are performed through human feedback/AI feedback.

 % As the model updates, the evaluation model may encounter over Out-of-Distribution (OOD) issues~\cite{gao2023scaling}, leading to validity bias. To correct the validity bias issue, 

 % \textbf{The inner loop} carries out model iteration updates, enhancing model capabilities through instruction responses and preference evaluations, while simultaneously accumulating preference data. \textbf{The outer loop} subjects the accumulated preference data to human and AI verification, with data showing significant discrepancies being used as feedback to update the model and correct biases.

 The Self-Reward~\cite{yuan2024self} method treats the model as an evaluator to score the generated responses. However, this process does not include human feedback, which can lead to out-of-distribution evaluation results. To incorporate human preferences, we use a scored dataset  from human feedback to perform supervised fine-tuning on the model.

% Specifically, we utilize a base model and minimize the supervised fine-tuning loss (Equation \ref{equ:sft_loss}) to obtain the evaluation model $\mathcal{M}_{t_b,s_b}^{SFT}$.
% \begin{align}
%     \label{equ:sft_loss}
%     \mathcal{L}_{SFT}(\theta)=-\mathbb{E}_{(x,y)\sim \mathcal{D}_s}[\log \mathcal{M}_{t_b,s_b}(y|x,\theta)]
% \end{align}
% where, $\theta$ denotes the learnable parameters in $\mathcal{M}_{t_b,s_b}$, $\mathcal{D}_s$ denotes the text-to-text scored instruction datasets, which means every response in the datasets has the tokens ``Scores:" in the end. Example shows in Appendix \ref{appendix:reward-correction}.

\begin{figure}[t]
    \centering
    \includegraphics[width=0.8\linewidth]{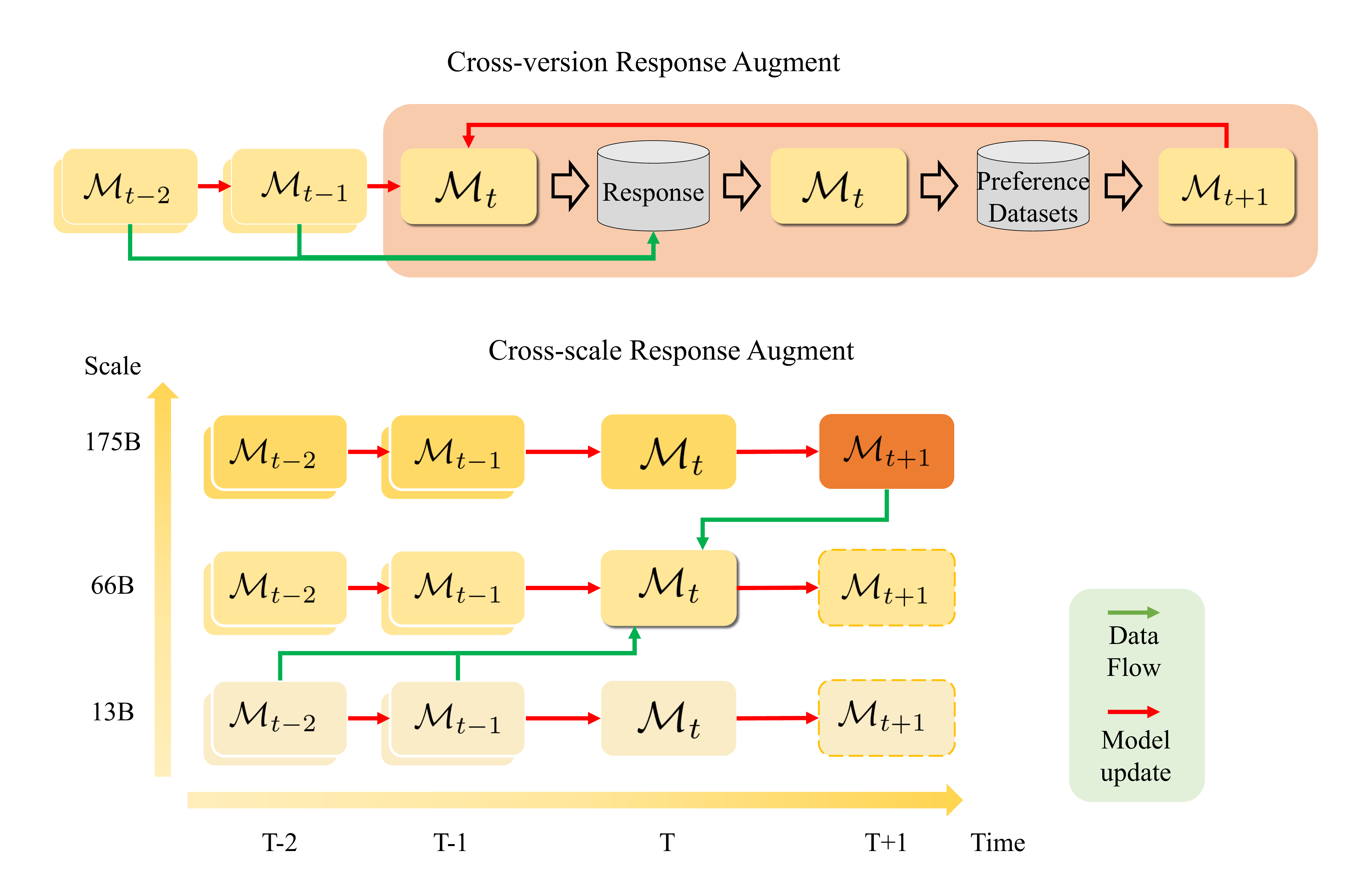}
    \caption{Model Matrix Instructions Construction. For cross-version augment, the model utilizes inferences from the older version of the model as candidate negative responses. For cross-scale augment, the model utilizes inferences from a smaller model as candidate negative responses. Meanwhile, the latest and largest model's inferences are used as candidate positive responses.}
    \label{fig:lpo-mmic}
\end{figure}

Specifically, we first utilize a base model and train a evaluator  $\mathcal{M}_{base}^{SFT}$ according to the  LLM-as-a-Judge~\cite{zheng2024judging}  manner. Furthermore, we utilize the evaluation model to evaluate the inference results of the base model and expand the response dataset constructed by the model matrix. The expanded instruction dataset is as follows:

\begin{align}
    \Tilde{\mathcal{D}}_{I}^w=&  \mathcal{D}_{I}^w  \cup \notag \\ & \{(x,y)| y \sim \mathcal{M}_{base}, \delta(\mathcal{M}_{base}^{SFT}(x,y)) >\tau \}
\end{align}
\begin{align}
    \Tilde{\mathcal{D}}_{I}^l =& \mathcal{D}_{I}^l \cup \notag \\ & \{(x,y)| y \sim \mathcal{M}_{base}, \delta(\mathcal{M}_{base}^{SFT}(x,y)) \leq \tau \}
\end{align}
where,  $\tau=\mathbb{E}_{(x,y)\sim \mathcal{M}_{base}}[\delta(\mathcal{M}_{base}^{SFT}(x,y))]$, $\mathcal{M}_{base}=\mathcal{M}_{t_b,s_b}$, $\delta(\cdot)$ represents the operation of extracting evaluation score from the responses.

Finally, we use the expanded instruction dataset to construct the preference dataset (Equation \ref{equ:pre-data}) for DPO training.
\begin{align}
\label{equ:pre-data}
    \mathcal{D}_{pre}=\{(x,y_w,y_l) | y_w\sim\Tilde{\mathcal{D}}_{I}^w,  y_l\sim \Tilde{\mathcal{D}}_{I}^l \}
\end{align}

\subsection{Mini-Batches Iterative DPO}\label{sec:training-strategies}
 Drawing inspiration from the deep learning concept of mini-batches,we  evenly divides the preference dataset into $T$ mini-batches. Each DPO training session only processes a single mini-batch and continuously updates the  responses of the reference model, aiming to fully exploit the potential of the preference dataset.
 
 Meanwhile, We found that during the optimization process using the original DPO loss (Equation \ref{equ:dpo-loss}), negative responses always had a dominant advantage. To balance the optimization process of positive and negative responses, we propose the Dual Clip Reward Loss(Equation \ref{equ:clip-loss}).
 % \textbf{$\mathcal{L}_{dual-clip}$}  ~\ref{equ:clip-loss} is designed to prevent either positive or negative responses from dominating the direction of optimization during the process. By separating the rewards for positive and negative responses and individually setting margins, i.e. , ($\gamma_w$, $\gamma_l$) .  $\mathcal{L}_{dual-clip}$ aims to balance the levels of optimization between positive and negative responses compared to the original DPO Loss ~\ref{equ:dpo-loss}, where $\sigma(x):=1/(1+\exp({-x}))$.
\begin{align}
    \mathcal{L}_{dual-clip} &= \mathbb{E}_{(x,y_w,y_l) \sim \mathcal{D}_{pre}}\notag \\
    &[\max(0,\gamma_w-\beta\log \frac{\pi_{\theta}(y_w|x)}{\pi_{ref}(y_w|x)})  \notag \\
+ &\max(0,\gamma_l+\beta\log\frac{\pi_{\theta}(y_l|x)}{\pi_{ref}(y_l|x)})] \label{equ:clip-loss}
\end{align}
\begin{align}
        \mathcal{L}_{DPO} &= \mathbb{E}_{(x,y_w,y_l) \sim \mathcal{D}_{pre}}\notag \\&
        [\log\sigma(\beta\log \frac{\pi_{\theta}(y_w|x)}{\pi_{ref}(y_w|x)}-\beta\log \frac{\pi_{\theta}(y_l|x)}{\pi_{ref}(y_l|x)})]\label{equ:dpo-loss}
\end{align}
where, $\pi_{\theta}$ is the model we aim to optimize, $\pi_{ref}$ is the reference model, and $\sigma(x) := 1/(1+\exp{(-x)})$. $\beta$ is a  predefined hyper-parameter, $\gamma_w$ and $\gamma_l$ are the clip margin to balance the optimization process of positive and negative responses. Hyper-parameter setting can refer to Appendix \ref{appendix:training-detail}

\section{Experiments}
\subsection{Experimental Setup}
\subsubsection{Base Model}
We utilize transformer-based models with a LLaMa-like~\cite{touvron2023llama} architecture as our base model, which has 66B parameters ($\mathcal{M}_{base}$). Building upon this base model, we conduct further experiments to align the model with human preferences. Additionally, our model matrix also includes different  versions and sizes of 13B, 66B and 175B models. To simplify the description, we denote these different sizes model as \texttt{TSO-M-13B}, \texttt{TSO-M-66B}, \texttt{TSO-M-175B}. For more details about the model architecture, please refer to Appendix~\ref{appendix:model-detail}.
\subsubsection{Datasets}\label{sec:data}

We've compiled a collection of 30,000 questions from both public datasets and our own sources. The public and custom datasets we constructed  are summarized as follows:

\begin{itemize}
    \item \textbf{HH-RLH}F~\cite{bai2022training} Dialogues between a human and an AI assistant are structured such that each conversation includes two potential responses from the AI - one that is preferred and another that is not, as judged by a human annotator. Preferences are determined based on the informativeness and honesty of the response for aiding tasks, and the safety of the response for non-harmful tasks.
    \item \textbf{Reddit TL;DR}~\cite{volske-etal-2017-tl} Content from Reddit along with condensed summaries of each post.
    \item \textbf{TSO-D} encompasses a broad range of themes (including social sciences, natural sciences and so on) and diverse task types (such as knowledge-based Q\&A, summarization and so on). 
\end{itemize}

After assembling this 30,000-prompt dataset, we engaged human annotators to provide answers for each question.

 Our model matrix comprised aligned models including the \texttt{TSO-M-66B} base model, prior versions of \texttt{TSO-M-66B}, every version of \texttt{TSO-M-13B}, and the latest version of \texttt{TSO-M-175B}. Inference  for the 30K prompts dataset from models other than the base model were pre-stored. For the base model, subsequent to each training iteration, it re-inferred the prompts dataset to update the responses. Drawing on the methodologies established in Section~\ref{sec:lpo}, we designated the responses  from the Human Datasets and the latest version of \texttt{TSO-M-175B} as positive responses, while those from earlier versions of \texttt{TSO-M-66B} and all versions of \texttt{TSO-M-13B} were deemed negative responses, thereby constructing a preference dataset featuring multi-model responses. Each category of positive response comprises an equal 50\% from the human datasets and inference results from \texttt{TSO-M-175B}, whereas negative responses are uniformly sampled from the respective model inference. Conversely, consistent with prior research ~\cite{yuan2024self}, the outputs post-inference, scoring, and selection by the latest base model form the preference dataset for the baseline responses. Ultimately, following uniform sampling, the multi-model and base model preference datasets each account for 50\% of the total training dataset, with the aggregate dataset encompassing 30K preference pairs.

More information about  the distribution of prompts is detailed in Appendix~\ref{appendix:data-distribution}.
  
  % \textbf{Human Datasets} Upon aggregating the aforesaid 30K-prompt dataset, we recruited human annotators to answer each question. Before the commencement of the annotation process, we defined rigorous standards for the answers, emphasizing accuracy, appropriateness, and non-maleficence. Upon completion of the annotations, cross-validation among annotators was performed to verify adherence to the established criteria.

  % \textbf{Model Generated Datasets} (target model, other version or scale models)

  % \textbf{Preference Pair Construction} 

% \small
% \setlength{\tabcolsep}{5pt}
\begin{table*}[t]
\centering

\renewcommand\arraystretch{1.2}
\begin{tabular}{ccccc}
\hline
{ Method}     & { AlignBench} & { MT-Bench} & { AlpacaEval-v2} & {Arena-Hard} \\ \hline
{ $\mathcal{M}_{base}$} & { 6.08}       & { 6.99}     & { 17.51\%}       & { 15.40\%}      \\ 
\rowcolor{gray!20} { DPO}        & { 6.40}       & { 7.49}     & { 27.84\%}       & { 19.40\%}      \\
{ RSO}        & { 6.17}       & { 7.35}     & { 24.63\%}       & { 16.00\%}      \\
\rowcolor{gray!20}{ IPO}        & { 6.17}       & { 7.39}     & { 26.21\%}       & { 17.40\%}      \\
{ cDPO}       & { 6.35}       & { 7.28}     & { 22.94\%}       & { 14.70\%}      \\
% \rowcolor{gray!20} { PPO}        & 6.18      &    7.20 &  25.83 &      14.90                        \\

\rowcolor{gray!20}{ TSO-1}      & { 6.43}       & { 7.51}     & { 21.65\%}       & { 21.10\%}       \\
{ TSO-2}      & { 6.74}       & { 7.49}     & { 26.35\%}       & { 26.90\%}       \\
\rowcolor{gray!20}{ TSO-3}      & { \textbf{6.96}\textcolor{red}{ (+0.88)}}      & { \textbf{7.55}\textcolor{red}{ (+0.56)}}     & { \textbf{29.47\%}\textcolor{red}{ (+11.96\%)} }      & { \textbf{30.80\%}} \textcolor{red}{ (+15.4\%)}       \\ \hline

\end{tabular}
\caption{ Results from multiple alignment evaluation sets are presented. All methods commence training with $\mathcal{M}_{base}$. Comparative experiments utilize a 30K-entry Single-Model response preference dataset\protect\footnotemark. These experiments undergo three rounds of iteration, where each stage of TSO utilizes the multi-model response preference dataset constructed as detailed in Section~\ref{sec:data}. TSO-1 denotes the initial stage of training employing the TSO method, followed sequentially by further stages.
}
\label{tab:main-result}
\end{table*}
\footnotetext{For the previously collected 30K prompts, we used human response as the positive responses and the latest version \texttt{TSO-M-175B} as the negative responses, forming a Single-Model preference dataset.}

\subsubsection{Evaluation Benchmark}
To evaluate the human preference alignment effect of TSO, we employed both publicly available and proprietary automatic evaluation datasets, including AlignBench ~\cite{liu2023alignbench}, MT-Bench ~\cite{zheng2024judging}, AlpacaEval-v2 ~\cite{li2023alpacaeval}, Arena-Hard ~\cite{li2024crowdsourced}, and our proprietary evaluation dataset TSO-Self-Bench-2K. All models subject to evaluation adhere to the same evaluation hyperparameters settings, ensuring the fairness and reproducibility of the results. More detailed information regarding the evaluation sets and evaluation hyperparameters settings are provided in Appendices ~\ref{sec:eval-hyperparams}.

\subsubsection{Training Detail}

The specifics of the models within the model matrix are delineated in Appendix~\ref{appendix:model-detail}. The utilized training hyperparameters are detailed as follows: for TSO training, the Adam optimizer ~\cite{kingma2014adam} is deployed, configured with a learning rate of $1 \times 10^{-6}$, a weight decay rate of 0.05, Adam $\beta_1$ of 0.9, and Adam $\beta_2$ of 0.95. The strategy for learning rate adjustment employs a cosine function with a warm-up mechanism, where the learning rate decreases to a minimum of zero. For the preference dataset, a batch size of 256 is used, with each TSO training cycle consisting of two epochs. Following the original DPO setup ~\cite{rafailov2024direct}, both $\mathcal{L}_{DPO}$ and $\mathcal{L}_{dual-clip}$ set $\beta$ at 0.1. The number of mini-batches $T$ is 3, and the total number of iterations $N$ is 3. Training of the \texttt{TSO-M-66B} DPO utilizes 64 Nvidia A800 GPUs, each with 80GB, processing an average of approximately 20 samples per second. Additional training specifics are provided in Appendix~\ref{appendix:training-detail}.

\subsection{Main Result}\label{sec:main-result}

Utilizing the 66B base model ($\mathcal{M}_{base}$), we conduct a three-stage TSO optimization. At each stage, we deploy a total of 30K preference pair from a multi-model response preference dataset, supplemented by preference data refined by the model itself. To ensure consistency in data volume across comparative experiments such as DPO and IPO, we employ 30K preference pair from a single-model response preference dataset across three iterations. Results from multiple alignment evaluation sets are displayed in Table ~\ref{tab:main-result}.

Compared to $\mathcal{M}_{base}$, TSO-3 exhibited improvements of \textbf{0.88} and \textbf{0.56} on AlignBench and MT-Bench,  on AlpacaEval-v2, the length-controlled win rate increased by \textbf{11.96\%}, and on Arena-Hard, it increased by \textbf{15.4\%}, respectively, indicating substantial effectiveness over the traditional DPO method.

Furthermore, while the DPO method demonstrates effectiveness and stability across various evaluation sets, TSO-2 has consistently outperformed DPO. TSO-3 further enhances the foundation established by TSO-2, showing no signs of performance deceleration. Specifically, on AlignBench, the progression from TSO-1 to TSO-2 result in an increase of \textbf{0.29}, and the advance from TSO-2 to TSO-3 yield an additional \textbf{0.22}. This suggests that the diversity of negative samples derived from the multi-model response preference dataset, coupled with self-corrections by the model, benefits  Mini-Batches Iterative DPO training.

\textbf{Negative Response Distribution}
To investigate the relationship between the negative response distribution engendered by various models and the base model, we streamlined the data generation process. We employed solely human and the latest version \texttt{TSO-M-175B} generated responses as positive inputs, and responses from a singular, weaker model as negatives, executing a single round of TSO training. Results are detailed in Table ~\ref{tab:model-alignbench} and Table ~\ref{tab:neg-distribution}.
\begin{table*}[htb]
\centering

\begin{tabular}{ccccccccc}

\hline
Model      & $\mathcal{M}_1$ & $\mathcal{M}_2$ & $\mathcal{M}_3$ & $\mathcal{M}_4$ ($\mathcal{M}_{base}$) & $\mathcal{M}_5$ & $\mathcal{M}_6$ & $\mathcal{M}_7$ & $\mathcal{M}_8$ \\
AlignBench & 5.63                            & 5.64                            & 6.08                            & 6.21                            & 6.28                            & 6.43                            & 6.92                            & 7.13                            \\ \hline
\end{tabular}
\caption{The scores of eight models in the model matrix on Alignbench increase from left to right.}
\label{tab:model-alignbench}
\end{table*}

It is important to note that utilizing responses from the $\mathcal{M}_{base}$
  itself as negative responses induces a phenomenon of reverse alignment, resulting in a significant deterioration in alignment performance. This decline is attributed to the high correlation between the distribution of negative responses and the base model. In DPO, the model is required to negate all its own answers, thereby disrupting the established alignment and significantly reducing model performance.Additionally, optimal alignment performance is achieved when engaging models slightly superior or inferior to the base model as sources of negative responses. Responses from overly proficient or inadequate models fail to facilitate improvements in alignment performance.
\begin{table}[htb]
\centering
\begin{tabular}{ccc}
\hline
POS-SRC                          & NEG-SRC                                 & TSO-Self-Bench-2K                 \\ \hline
                                 & $\mathcal{M}_1$                         & 4.09                         \\
                                 & $\mathcal{M}_2$                         & 4.09                         \\
                                 & $\mathcal{M}_3$                         & 4.12(\textcolor{red}{+0.26})                         \\
                                 & \cellcolor{gray!20}$\mathcal{M}_4(\mathcal{M}_{base})$& \cellcolor{gray!20}3.90 (\textcolor{blue}{+0.3})  \\
                                 & $\mathcal{M}_5$                         & 4.09                         \\
                                 & $\mathcal{M}_6$                         & 4.12\textcolor{red}{+0.26}                         \\
                                 & $\mathcal{M}_7$                         & 4.10                         \\
\multirow{-8}{*}{$\mathcal{M}_w$} & $\mathcal{M}_8$                         & 4.05                         \\ \hline
\end{tabular}
\caption{Negative Response Distribution: POS-SRC denotes models producing positive responses, encompassing both human-derived and the latest version \texttt{TSO-M-175B} generated responses (Equation \ref{equ:M_w}). NEG-SRC denotes models yielding negative responses, which include diverse sizes and versions of the models, systematically arranged in ascending order of their AlignBench scores.\protect\footnotemark }
\label{tab:neg-distribution}
\end{table}
\subsection{Ablation}
In Section~\ref{sec:lpo}, we constructed multi-model preference data through a model matrix, significantly expanding the diversity of the training data. This was further refined through human and AI feedback for validity bias correction. Ultimately, we employed training strategies using Mini-Batches Iterative DPO and Dual Clip Reward Loss to update model responses and balance the optimization of positive and negative responses. To validate the effectiveness of our method, we posed the following questions and conducted experiments to address them systematically: \textbf{\textit{Q1).}}  Does the multi-model  preference dataset  help improve the alignment effect? \textbf{\textit{Q2).}} Has $\mathcal{L}_{dual-clip}$ shown improvement compared to $\mathcal{L}_{DPO}$? \textbf{\textit{Q3).}} Does the design of Mini-Batches Iterative DPO strategy have an effect? \textbf{\textit{Q4).}} How is the model's evaluation and correction capability?

\subsubsection{\textit{Q1.} }
To ensure fairness in comparative evaluations, we have refined the Self-Reward methodology ~\cite{yuan2024self}, hereafter referred to as Self-Reward\dag. Unlike the original method, we employ an external reward model (Qwen2 72B~\cite{yang2024qwen2technicalreport}) rather than the model itself to assess multiple generated responses from various dimensions, such as comprehension, conciseness, factuality, and logic. We select the response with the highest average score as the positive and that with the lowest score as the negative to generate preference data for DPO training. Similar to TSO, both methods implement a three-stages iterative learning process, continuously updating the reference model.

\begin{table}[htb]

\centering
\begin{tabular}{
>{\columncolor[HTML]{FFFFFF}}c 
>{\columncolor[HTML]{FFFFFF}}c 
>{\columncolor[HTML]{FFFFFF}}c }
\hline
{ Stage}                      & { Method}       & { AlignBench} \\ \hline
{ $\mathcal{M}_{base}$}                 & { -}            & { 6.08}       \\ \hline
                          & { TSO}          & { \textbf{6.43}}       \\
\multirow{-2}{*}{Stage 1}  & { Self-Reward\dag} & 6.22                              \\ \hline
                          & { TSO}          & { \textbf{6.74}}       \\
\multirow{-2}{*}{Stage 2} & { Self-Reward\dag} & 6.12                              \\ \hline
                          & { TSO}          & { \textbf{6.96}}       \\
\multirow{-2}{*}{Stage 3} & { Self-Reward\dag} & 6.21                              \\ \hline

\end{tabular}
\caption{Self-Reward\dag  vs  TSO.}
\label{tab:self-r_vs_lpo}
\end{table}

Upon analysis of Table ~\ref{tab:self-r_vs_lpo}, it is evident that within the AlignBench evaluation set, the enhancements at each stage of TSO are more pronounced compared to those of the modified Self-Reward method. Interestingly, the modified Self-Reward method exhibits a decline in model performance between stage 1 and stage 2. This observation suggests that preference data generated solely through leveraging the model’s own inference capabilities inherently possesses flaws characteristic of the base model. Such flawed preference data proves non-conducive to aligning the model with genuine human preferences in later stages, i.e. , as iterations progress, the distribution of model-generated preference data diverges from the authentic distribution of human preferences. However, TSO enhances the diversity of negative responses effectively by incorporating feedback from various temporal versions and different model scales. This enables the model to circumvent several types of deficiencies, thereby generating a more robust preference dataset conducive to iterative DPO training.
\subsubsection{\textit{Q2.}}

{
% \small
% \setlength{\tabcolsep}{3pt}
\begin{table*}[htbp]
\centering

\begin{tabular}{ccccc}

\hline
{ Method}     & { Align-Bench} & { MT-Bench} & { AlpacaEval-v2} & {Arena-Hard} \\ \hline
$\mathcal{M}_{base}$                                                         & { 6.08}        & { 6.99}     & { 17.51}       & { 15.40}      \\
{ $\mathcal{L}_{DPO}$-Single-Model-10K}       & 6.36                               & 7.44                            & { 25.74}       & { 20.91}      \\
{ $\mathcal{L}_{dual-clip}$-Single-Model-10K} & \textbf{6.39}                           & \textbf{7.55}                            & { \textbf{28.87}}       & { \textbf{23.65}}      \\ \hline

\end{tabular}
\caption{$\mathcal{L}_{dual-clip}$ vs $  \mathcal{L}_{DPO}$.}
\label{tabel:clip-loss}
\end{table*}

}

\begin{figure}[bhtp]
    \centering
    \includegraphics[width=0.8\linewidth]{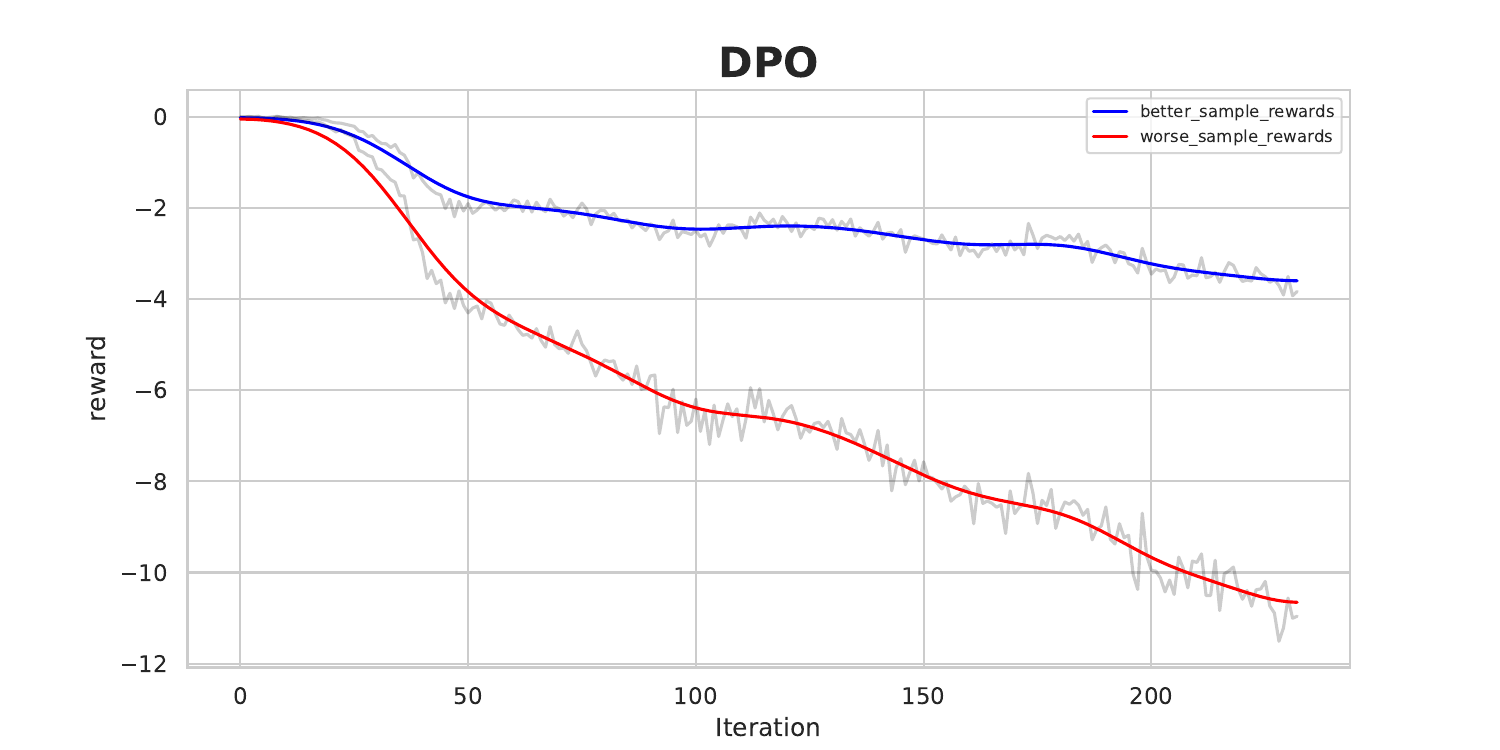}
    \caption{\textbf{DPO} stands for using  $\mathcal{L}_{DPO}$ . The \textcolor{blue}{blue} line signifies the rewards obtained from positive responses, i.e. , $\beta\log \frac{\pi_{\theta}(y_w|x)}{\pi_{ref}(y_w|x)}$ , while the \textcolor{red}{red} line indicates the rewards obtained from negative responses, i.e. , $\beta\log \frac{\pi_{\theta}(y_l|x)}{\pi_{ref}(y_l|x)}$.
    }
    \label{fig:reward-dpo}
\end{figure}

 To ascertain the effectiveness of $\mathcal{L}_{dual-clip}$,  we executed an ablation study by comparing its performance with that of the original DPO loss across multiple alignment evaluation sets, as delineated in Section~\ref{sec:lpo}. To streamline the training process, we extracted a subset of 10K preference data points and a single model response as the negative response from the initial 30K data points for training. The experimental results are delineated in Table~\ref{tabel:clip-loss}. The findings demonstrate that, relative to the original DPO loss, our  $\mathcal{L}_{dual-clip}$ achieves superior outcomes on several publicly available alignment evaluation sets, under identical base models and datasets.

\begin{figure}[bhtp]
    \centering
    \includegraphics[width=0.8\linewidth]{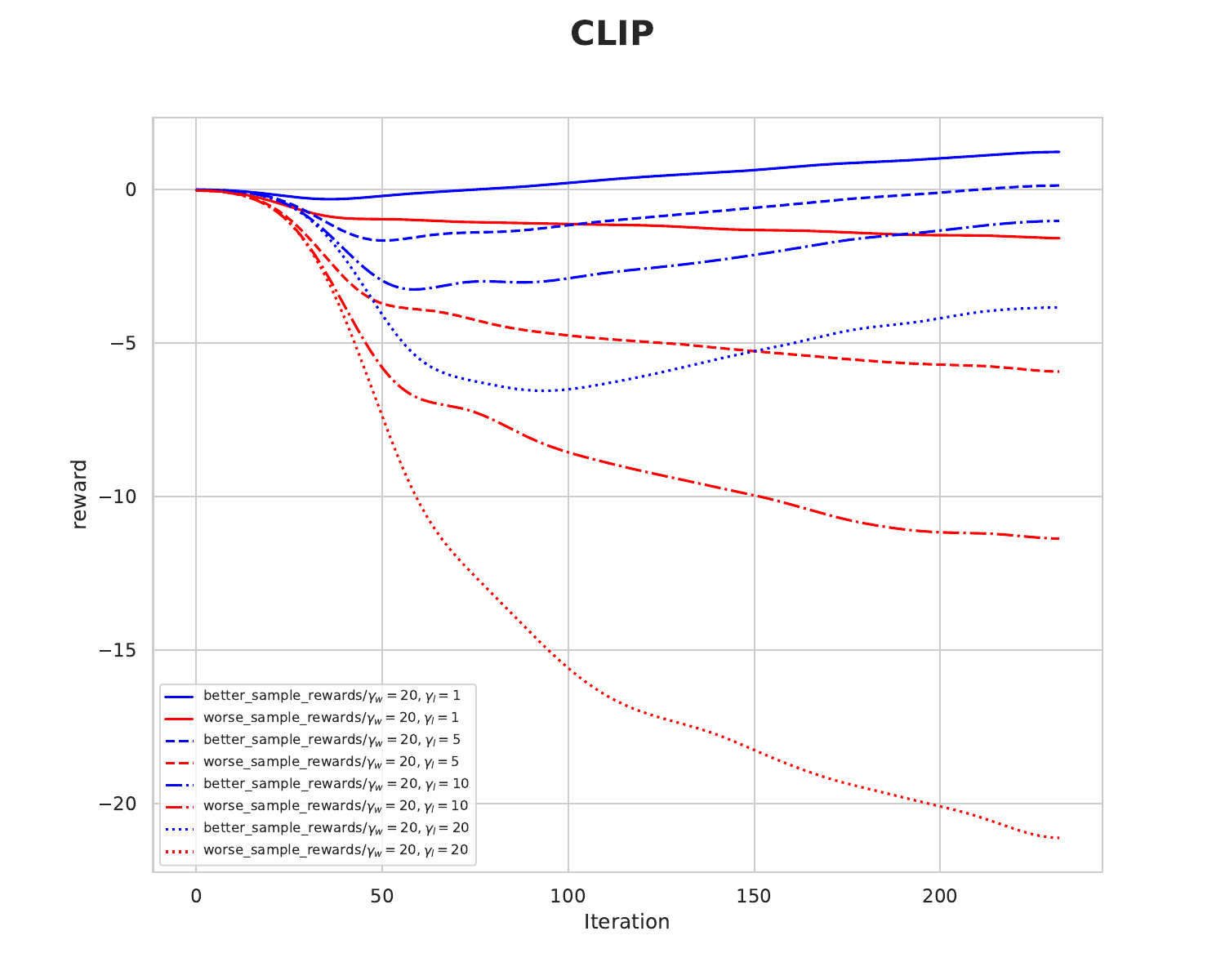}
    \caption{\textbf{CLIP} represents using $\mathcal{L}_{dual-clip}$.}
    \label{fig:reward-clip}
\end{figure}

Reward Curve \& Explanation: we plot the changes for positive and negative responses' rewards during the optimization process, as shown in Figure~\ref{fig:reward-dpo} and Figure~\ref{fig:reward-clip}.

 It can be observed that  using the $\mathcal{L}_{DPO}$, due to the coupling of the positive and negative response losses, the negative responses dominate throughout the optimization process, leading to a decrease in the rewards for positive responses. Compared to the $\mathcal{L}_{DPO}$, $\mathcal{L}_{dual-clip}$ shows similar behavior to the $\mathcal{L}_{DPO}$ in the early stages of optimization because neither positive nor negative responses are truncated during the initial phase. In the middle and later stages of optimization, the reward for positive responses increases, while the reward for negative responses decrease. This is due to the smaller margin for negative responses. The loss from negative responses begins to be truncated and thus ceases to contribute to the optimization process, while the effect of negative response optimization on positive responses decreases, resulting in an upward trend in the rewards for positive responses. This demonstrates that $\mathcal{L}_{dual-clip}$ can balance the optimization processes for positive and negative responses.

Meanwhile, as  $\gamma_l $ increases, the final rewards obtained by both positive and negative samples are generally reduced, and the absolute value of the reward margin gradually increases. Moreover, since $\mathcal{L}_{dual-clip}$ avoids the coupling of optimization between positive and negative samples, as $\gamma_l$ increases, the reward margin of \textbf{CLIP} will gradually surpass that of DPO.

\footnotetext{The score of ($\mathcal{M}_{base}$) on TSO-Self-Bench-2K is 3.87.}
{
% \small
% \setlength{\tabcolsep}{3pt}
\begin{table*}[htpb]
\centering

\begin{tabular}{ccccc}
\hline
Method     & { Align-Bench} & { MT-Bench} & { AlpacaEval-v2} & {Arena-Hard}   \\ \hline
$\mathcal{M}_{base}$                     & { 6.08}       & { 6.99}     & { 17.51}       & { 15.40}      \\
{ DPO-Single-Model-30K} & 6.36                             & 7.44                           & { 25.74}       & { 20.91}      \\
{DPO-Single-Model-30K-MiniBatch-1} & 6.23                            & \textbf{7.60}                       & { 26.00}       & 21.59                             \\
{ DPO-Single-Model-30K-MiniBatch-2} & \textbf{6.42}                              & 7.38                          & \textbf{ 28.52}       & 21.45                             \\
{DPO-Single-Model-30K-MiniBatch-3} & 6.40                             & 7.54                          & { 26.89}      & \textbf{22.27  }                           \\ \hline

\end{tabular}
\caption{Mini-Batches Iterative DPO.}
\label{tab:iterative}
\end{table*}
}

\subsubsection{\textit{Q3.}}
 In the methodology outlined in Section~\ref{sec:lpo}, we segmented the dataset, updating the reference model's probability response to immediately reflect the target model’s after each update and learning cycle within a single minibatch. However, does it enable better learning from the preference dataset? To investigate this, we intend to evenly split the original dataset (30K) into three distinct segments, with each DPO training session handling only mini-batches of 10K. To streamline the training process, responses from a single model were employed as negative responses. The comparative outcomes are presented in Table~\ref{tab:iterative}. The results indicate that the outcomes in the third stage excel over those of the initial DPO settings under equivalent data length. The rationale behind this phenomenon, from the perspectives of data utilization efficiency and gradient orientations, is elaborated in Appendix~\ref{appendix:iter-dpo}.

\subsubsection{\textit{Q4.}}
 To validate the correction capabilities of the model's scoring ability following human and AI feedback, we designed the experiment described below. Initially, the unmodified $\mathcal{M}_{base}$  directly scored the QA pairs generated by our model matrix, based on predefined criteria: factuality, conciseness, logic, and comprehension. Subsequently, we constructed a scoring dataset that included feedback from both human evaluators and AI, employed to train the base model through Supervised Fine-Tuning (SFT). The SFT model then reevaluates the QA pairs using these criteria to determine if the SFT effectively correctes out-of-distribution (OOD) samples. Experimental results suggest that following the implementation of SFT, the model effectively adjusted the scoring of OOD samples, encompassing both unfavorable and favorable cases. Further elaboration is provided in Appendix~\ref{appendix:reward-correction}.
% \begin{table}[bpth]
% \centering
% \caption{Reward Correction:{ $\mathcal{M}_{base}$-SFT} represents the model that has been fine-tuned with human scoring dataset based on { $\mathcal{M}_{base}$},the same as to $\mathcal{M}_{TSO-3}$-SFT }
% \begin{tabular}{cc}
% \hline
% { Model}          & { Accuracy(Preference)} \\ \hline
% { $\mathcal{M}_{base}$-SFT} & { 0.7015}          \\
% { $\mathcal{M}_{TSO-3}$-SFT} & { 0.6935}          \\ \hline
% \label{tab:reward-correct}
% \end{tabular}
% \end{table}

%  It can be observed that both the base model and the aligned TSO-3 model, after being fine-tuned with human scoring data, possess a certain capability to discriminate preferences, achieving an accuracy of around 70\%.This indicates that the model has a good capability for evaluation and correction.

% \section{Analysis and Discussion}
\section{Conclusion}
We introduce \textbf{TSO}, a direct preference optimization method based on multi-model responses. By generating diverse responses through a model matrix, this approach aims to augment the variety of the preference dataset. Additionally, it incorporates feedback from both humans and AI to enhance the model's evaluation and correction capabilities, and to mitigate the preference deficiencies arising from solely relying on self-model response adjustments. The training strategy employed includes Mini-Batches Iterative DPO and Dual Clip Reward Loss. Our experiments validate the effectiveness of TSO and various training strategies, and confirm that the improvements in alignment are due to the response diversity provided by the model matrix. Furthermore, we explore the relationship between the distribution of negative responses and the foundational model, providing insights into the construction of preference pairs.

\clearpage

\bibliographystyle{unsrtnat}
\bibliography{references}  

\clearpage

\appendix

\section{Data}\label{appendix:data-distribution}

\subsection{Data Distribution Analysis}
\begin{figure}[bpth]
    \centering
    \includegraphics[width=0.8\linewidth]{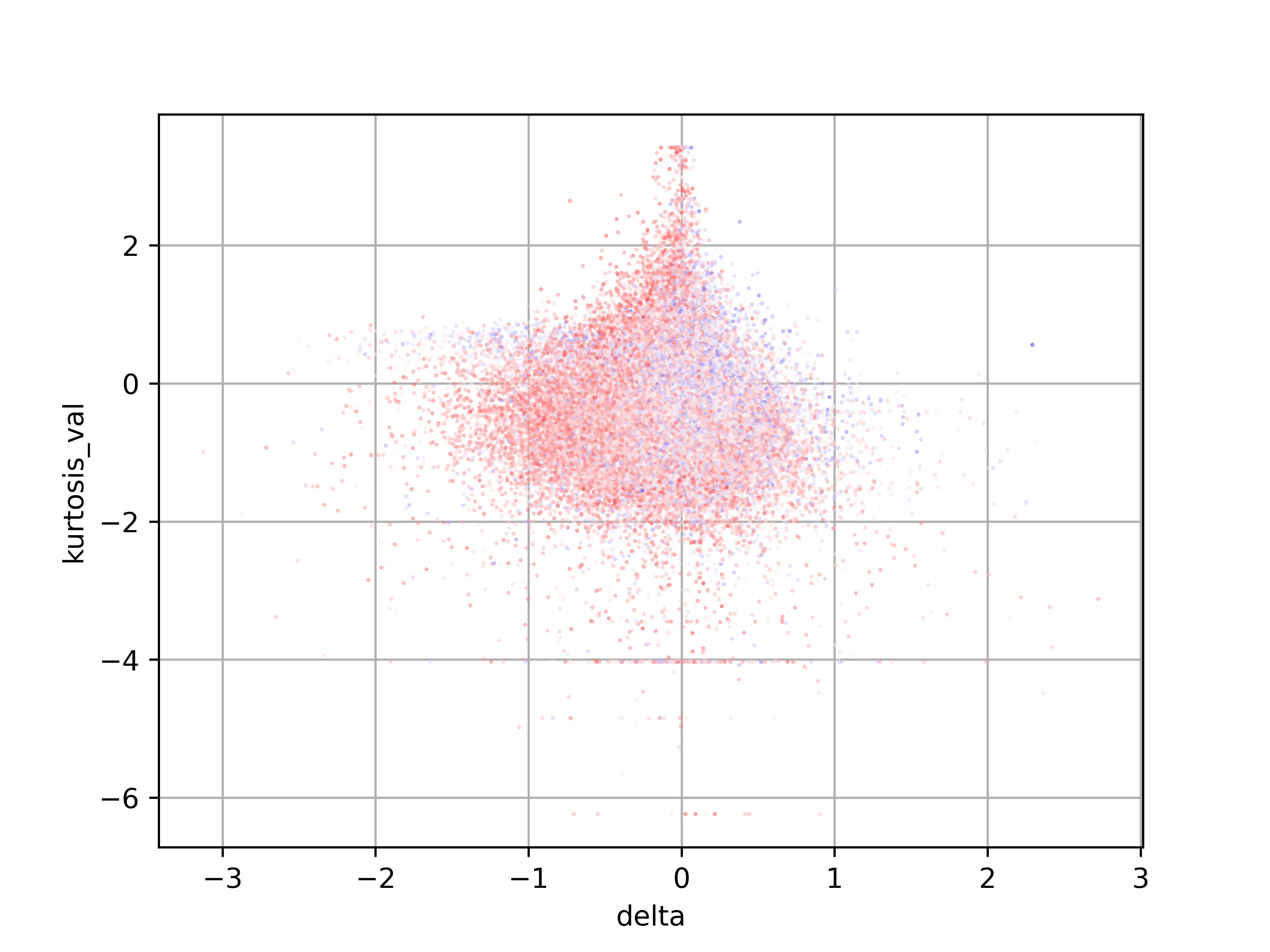}
    \caption{Data distribution. For each question, we generate 64 different answers and scores. Every point stands for a question.``delta'' represents the skewness of scores distribution, ``kurtosis\_val'' represents the normalized kurtosis of scores distribution.}
    \label{fig:data-dis}
\end{figure}
To understand the score distribution and patterns of various types of question-and-answer pairs collected during Section~\ref{sec:data}, we score multiple answers for each question and aggregate them based on the question, ultimately summarizing the data characteristics of the score distribution. This will help us gain a deeper understanding of the relationship between questions and their corresponding answers, thereby enhancing the accuracy and effectiveness of problem-solving.

\textbf{First}, we observed that a large amount of the data is concentrated on the negative half-axis of normalized kurtosis, indicating that the score distribution of most answers is relatively uniform with lower certainty. In other words, there is no obvious concentrated scoring area, reflecting the overall diversity of the answers. \textbf{Secondly}, we found that the data tends to cluster on the negative half-axis of skewness , suggesting that most answer distributions exhibit a characteristic of ``high scores concentrated, low scores long-tailed.'' This result indicates that among the higher scoring answers, there may exist some  ``severely low-quality'' answers. \textbf{Finally}, based on the variation in color intensity, we can observe that the color in the region around the origin is lighter, while the color away from the origin is darker, indicating that scores tend to be lower for normal distributions, whereas for cases with peaks or more uniform distributions, the scores are relatively higher.

\section{Details of the Model Architecture}\label{appendix:model-detail}

In our model matrix, we utilized three differently-sized models, including 13B, 66B, and 175B, all of which are based on the Llama2~\cite{touvron2023llama} architecture. Specific architectural details of the models are shown in Table~\ref{tab:model-arch}.All models in model matrix has been autoregressively pre-trained on several terabytes of  corpora and subsequently supervised fine-tuned  using a meticulously curated instructions datasets. 

\begin{table*}[bthp]
\centering

\begin{tabular}{ccccccccc}
\hline
Model & $d_{hidden}$ & $d_{FFN}$ & $N_{layer}$ & $N_{head}$ & $N_{voc}$ & $f_{act}$ & $T_{pos}$ & $T_{torch}$ \\ \hline
\texttt{TSO-M-13B}  & 5120         & 13824     & 40          & 40         & 80496     & SiLU& RoPE  & Float32   \\
\texttt{TSO-M-66B}  & 8192         & 22016     & 80          & 64         & 80496     & SiLU      & RoPE  & Float32    \\
\texttt{TSO-M-175B} & 12288        & 32768     & 96          & 96         & 128000    & SiLU      & ALiBi  & Float32   \\ \hline
\end{tabular}

\caption{Architectural details of models of various sizes in the model matrix. Here, $d_{hidden}$ represents the hidden dim of Transformer blocks.$d_{FFN}$ represents the hidden dim of the Feed-Forward Network.$N_{layer}$ represents the number of Transformer blocks.$N_{head}$ represents the head num of the Multi-Head Attentions (MHA).$N_{voc}$ represents the size of the vocabulary.$f_{act}$ represent the activation function in FFN.$T_{pos}$ represents the type of position embedding.$T_{torch}$ represents the data type of the model parameter tensors. SiLU is presented in ~\cite{elfwing2018sigmoid}. RoPE is presented in ~\cite{su2024roformer}. ALiBi is presented in ~\cite{press2021train}.}
\label{tab:model-arch}
\end{table*}

\section{Details of Training and Evaluation}\label{appendix:training-detail}

\subsection{Hardware and Software}
We conducted our training using eight machines equipped with Intel (R) Xeon (R) Platinum 8468 processors featuring 40 cores and 500 GiB, each machine outfitted with 8 Nvidia 80GB A800 GPUs. The operating system used is Ubuntu 20.04.6. Pytorch's version is 2.1.0a0+gitda1ccca.

\subsection{Training}
    \subsubsection{Method Introduction}
    In the comparative experiment ~\ref{sec:main-result}, we employed the DPO, IPO, cDPO, RSO, and PPO methods. The details for \textbf{TSO} and DPO have already been extensively discussed in Section~\ref{sec:training-strategies}. We now supplement the loss functions used for IPO and cDPO. The definitions for the IPO and cDPO losses are presented in Equation ~\ref{equ:ipo} and Equation ~\ref{equ:cdpo}.
    \begin{align}
        \mathcal{L}_{IPO}=\mathbb{E}_{(x,y_w,y_l)\sim \mathcal{D}}[(h(x,y_w,y_l)-\frac{1}{2\tau})^2]
        \label{equ:ipo}
    \end{align}

    \begin{align}
        \mathcal{L}_{cDPO}& =\mathbb{E}_{(x,y_w,y_l)\sim \mathcal{D}}[(1-\epsilon)\log\sigma(h(x,y_w,y_l))+ \notag \\   
        & \epsilon\log\sigma(-h(x,y_w,y_l))]
        \label{equ:cdpo}
    \end{align}

    where $h(x,y_w,y_l)=\beta\log\frac{p_{\theta}(y_w|x)}{p_{ref}(y_w|x)}-\beta\log\frac{p_{\theta}(y_l|x)}{p_{ref}(y_l|x)}$,$\tau$ in $\mathcal{L}_{IPO}$ and $\epsilon$ in $\mathcal{L}_{cDPO}$ is set particularly.

    Furthermore, PPO actor's training objective is defined in Equation ~\ref{equ:ppo}.

    \begin{align}
        \mathcal{L}_{PPO-actor}&=\mathbb{E}_{x\sim\rho, y_t\sim \pi_{old}(y_t|x,y_{<t})}\notag \\
        &[\hat{A}_t\min(r_t(\theta),clip(r_t(\theta),1-\epsilon,1+\epsilon))]
        \label{equ:ppo}
    \end{align}
    where $r_t(\theta)=\frac{p_{\theta}(y_t|x,y_{<t})}{p_{\pi_{old}}(y_t|x,y_{<t})}$ ,$\hat{A}_t$ represents the Generalized Advantage Estimation (GAE)~\cite{schulman2015high}. The ultimate goal is to maximize the rewards obtained from the sequence of answered questions, as defined in Equation ~\ref{equ:reward}.
    \begin{align}
        \max \mathbb{E}_{x\sim \rho ,y\sim\pi_{\theta}}RM(x,y)-\lambda_{KL}D_{KL}(\pi_{\theta}||\pi_{ref})
        \label{equ:reward}
    \end{align}
    where $RM$ represents reward model, $D_{KL}$represents Kullback–Leibler divergence. $\pi_{ref}$represents reference model.

    \subsubsection{Hyperparameters Setting}
    For the training of \textbf{TSO}, DPO, IPO, cDPO, we uniformly use the same experimental configuration. Initially, we set the learning rate to $1e-6$ and employ a cosine scheduler to facilitate the reduction of the learning rate. The constraint coefficient for weight L2 regularization is set at 0.05, and the gradient norm clipping threshold is set at 1.0.  Additionally, we use the Adam optimizer with parameters $\beta_1=0.9$, $\beta_2=0.95$, and $\epsilon=1e-8$. The random seed is fixed at 43.  The $\beta$ in Equation ~\ref{equ:clip-loss}, ~\ref{equ:dpo-loss}, ~\ref{equ:ipo}, ~\ref{equ:cdpo} is all set to 0.1. Batch size for all above methods is 256.
    
   For the clipping margins   $\gamma_w$ and $\gamma_l$ in $\mathcal{L}_{dual-clip} $of the TSO, we set them to 20 and 10 respectively. 
   
   For the $\tau$ in $\mathcal{L}_{IPO}$, we set it to 0.2, and  the $\epsilon$ in $\mathcal{L}_{IPO}$ is set to 0.3.

     For RSO, we first perform inference on each question eight times using $\mathcal{M}_{base}$ with different temperatures and random seeds to obtain eight distinct answers. Subsequently, we conduct rejection sampling according to ~\cite{liu2023statistical}. The experimental setup then follows the same process as DPO.

     For PPO, we set the $\lambda_{KL}$ in Equation~\ref{equ:reward} to 0.1. GAE parameters $\gamma$  is set to 1.0 and  $\lambda$ is set to 0.95.$\epsilon$ in Equation ~\ref{equ:ppo} is set to 0.2. The prompts in experience buffer is 128, and we use experience buffer 3 times for each sampling.
\subsection{Evaluation Hyperparameters}\label{sec:eval-hyperparams}

During the evaluation phase, we use a uniform inference setup for all models. Temperature is set to 0.7, $TOP_p$ for decoding is set to 0.9 and the maximum  input and output token length is set to 2048.
\subsection{Evaluation Benchmark}\label{sec:eval-bench}
\textbf{AlignBench} ~\cite{liu2023alignbench} functions as a comprehensive, multi-dimensional benchmark for assessing the alignment performance of Chinese large language models. AlignBench has implemented a human-involved data construction process to ensure the dynamic updating of evaluation data. It utilizes a multi-dimensional, rule-calibrated model evaluation approach (LLM-as-Judge) and integrates Chain-of-Thought to produce multi-dimensional analyses and a definitive comprehensive score for model responses, thereby enhancing the evaluation's reliability and interpretability. We deploy GPT4-0613 to conduct multi-faceted evaluations of the model-generated outcomes, ranging from \{1, 2, ..., 10\}. The evaluation dimensions of  AlignBench are displayed in Table~\ref{tab:eval-d}.

 \textbf{MT-Bench}~\cite{zheng2024judging} is a challenging multi-turn benchmark that measures the ability of large language models (LLMs) to engage in coherent, informative, and engaging conversations. It is designed to assess the conversation flow and instruction-following capabilities of LLMs, making it a valuable tool for evaluating their performance in understanding and responding to user queries. We use GPT4-0613 to conduct multi-dimensional evaluations on the multi-round results generated by the model, with the rating scale ranging from \{1, 2, ..., 10\}. The evaluation dimensions of  MT-Bench are displayed in Table~\ref{tab:eval-d}.

\begin{table*}[htbp]\
\centering
\begin{tabular}{c|c}
\hline
Benchmark  & Dimensions                                                                                                                                                                                                                                                   \\ \hline
AlignBench & \begin{tabular}[c]{@{}c@{}}Professional Competence, Chinese Comprehension, Basic Tasks, \\ Mathematical Calculation, Text Writing, Comprehensive Question-Answering, \\ Role-Playing, Logical Reasoning, Chinese Language, Chinese Reasoning\end{tabular} \\ \hline
MT-Bench   & \begin{tabular}[c]{@{}c@{}}Extraction, Humanities, Reasoning, Coding, Math, Roleplay, Writing, Stem\end{tabular}                                                                                                                                          \\ \hline
TSO-Self-Bench-2K & \begin{tabular}[c]{@{}c@{}}Correctness of Information, Comprehensibility, Targetedness, Safety, 
\\ Readability, Logicality, Self-awareness, Thoroughness, Creativity\end{tabular} \\ \hline

\end{tabular}
\caption{Evaluation benchmark dimensions.}
\label{tab:eval-d}
\end{table*}

 \textbf{AlpacaEval-v2} ~\cite{li2023alpacaeval}  is an automated tool for evaluating instruction-following language models against the AlpacaFarm dataset~\cite{dubois2024alpacafarm}. It stands out for its human-validated, high-quality assessments that are both cost-effective and rapid. We used GPT-4 Preview-1106 as the baseline and Auto-annotator, and reported the win rate of the model under test relative to the baseline in the experiment.

  \textbf{Arena-Hard}~\cite{li2024crowdsourced} serves as an automated evaluation tool for instruction-tuned large language models (LLMs). It encompasses 500 complex user queries. The Arena-Hard-Auto-v0.1 system employs GPT4-1106-preview as a judge to benchmark the models' responses against a default base model (GPT4-0314~\cite{achiam2023gpt}). We report the Length Controlled Win rate in the experiment section.

\textbf{TSO-Self-Bench-2K }represents a tailored alignment evaluation set , comprising 2206 meticulously curated questions across 13 themes, including humanities and mathematics. It bases its comparative assessments of model-generated results on GPT4-0613, with scores ranging from \{1, 2, 3, 4, 5\}, utilizing judging criteria that encompass correctness of information, comprehension, safety, readability, logicality, and other aspects, ultimately culminating in a composite average score across all dimensions.

\section{Ablation Supplementary }

\subsection{Mini-Batches Iterative DPO}\label{appendix:iter-dpo}

 \textbf{Gradient Explanation} Next, we will explain from the perspective of gradient magnitude why the introduction of mini-batches offers an advantage over the original DPO in terms of data information utilization efficiency.

\begin{figure}[bhtp]
    \centering
    \includegraphics[width=0.8\linewidth]{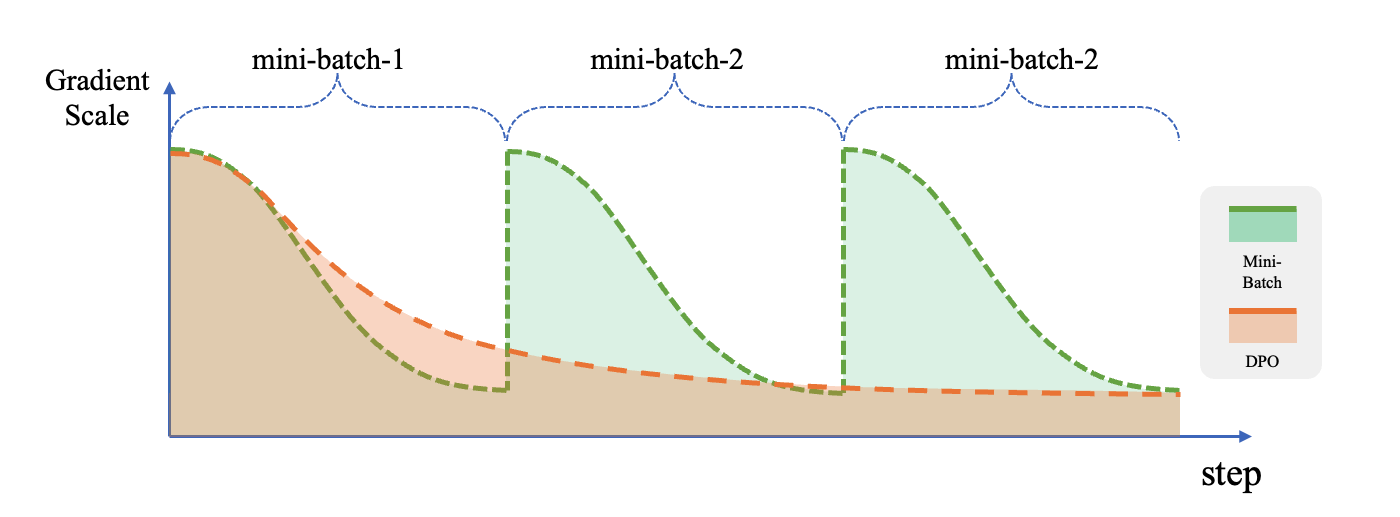}
    \caption{Mini-Batches Iterative DPO Gradient: The red curve represents the changes in $s$ for the original DPO during the training process. The green curve represents the changes in $s$ for DPO using mini-batches during the training. The dashed line indicates the moments when the reference model's parameters are updated to match those of the latest target model.}
    \label{fig:grad_mini-batches}
\end{figure}

The derivative of $\mathcal{L}_{DPO}$ with respect to the model training parameters $\theta$ can be obtained as follows:

\begin{align}
    \nabla_{\theta}\mathcal{L}_{DPO} &= \mathbb{E}_{(x,y_w,y_l) \sim \mathcal{D}}[- \notag \\ &s(\beta\nabla_{\theta}\log\pi_{\theta}(y_w|x) 
-\beta\nabla_{\theta}\log\pi_{\theta}(y_l|x))].
\label{equ:grad_dpo}
\end{align}
where,
\begin{equation}
s=\sigma(\beta\log\frac{\pi_{\theta}(y_w|x)}{\pi_{ref}(y_w|x)}-\beta\log\frac{\pi_{\theta}(y_l|x)}{\pi_{ref}(y_l|x)})
\end{equation}
 We have plotted the changes in the gradient scale $s$ during the training process for the original DPO and the DPO using three mini-batches, as illustrated in Figure~\ref{fig:grad_mini-batches}.
 It can be observed that for the original DPO, in the first third of the training, $s$ decreases rapidly to near zero. In the latter two-thirds, $s$ remains slightly above zero, with very small gradient magnitude. For the DPO employing mini-batches training, during the mini-batch-1 phase, its performance is similar to that of the original DPO. In the mini-batch-2 phase, due to the initial switch of the reference model to $\pi_{ref}=\pi_{\theta}, s = \sigma(0) = 0.5 $, the scale is reset, and the gradient magnitude remains relatively large. Hence, in the mini-batch-2 and mini-batch-3 phases, compared to the original DPO, the model better captures the human preference information carried in the dataset of these stages, resulting in higher data information utilization efficiency.

\subsection{Evaluation and Correction Capability}\label{appendix:reward-correction}

To validate the correction capability in the model’s
scoring ability after receiving human and AI’s feedback, we
designed the following experiment. Initially, $\mathcal{M}_{base}$, with-
out any modifications, directly scored the QA pairs, which is
generated from our model matrix, based on predefined criteria (factuality, conciseness, logic, and comprehension). Subsequently, we constructed a scoring dataset comprising both
human and AI feedback, which was used to train the base-
line model by Supervised Fine-Tuning. The fine-tuned model ($\mathcal{M}_{base-SFT}$)
then scored the QA pairs using the same criteria to assess
whether the supervised fine-tuning had corrected some of
the samples that were out-of-distribution (OOD). We list two examples in Table~\ref{tab:good-case} and ~\ref{tab:bad-case}. The former indicates that the $\mathcal{M}_{base-SFT}$ corrected the evaluation for good cases among OOD samples, while the latter shows that the $\mathcal{M}_{base-SFT}$ corrected the evaluation for bad cases among OOD samples.
\begin{table*}[bpth]
    \centering
    \begin{tabular}{{|p{18cm}|}}
    \hline
    
       [System]

We would like to request your feedback on the performance of the response of the assistant to the user question displayed below. In the feedback, I want you to rate the quality of the response in the following dimension to the given scoring rubric:This metric should be considered when the response involves the following scenarios, particularly when the user's intent included in the question is not easy to understand. Assess whether the text of the response understands the user's question, whether it can engage in a dialogue according to the user's intent, and whether it can understand or inquire to supplement information when the user's input is incomplete.

Score 2: The response is completely unrelated to the instruction, totally misunderstands the instruction, fails to identify any counterfactual or counterintuitive information in the instruction, or fails to inquire for additional information when the query is incomplete.

Score 4: Most key points in the response are unrelated to the instruction, and the response misses the main requirements of the instruction.

Score 6: Essentially meets the core requirements but misses some minor requirements, or mostly satisfies complex or multiple requirements with less than 30% unfulfilled.

Score 8: Meets the core requirements but fails to meet the minor requirements of the instruction, or has minor flaws in fulfilling complex requirements.

Score 10: Meets the core requirements, but either misses the minor requirements of the instruction or has minor flaws in fulfilling complex requirements.

You will be given an assistant's answer.

Please give feedback on the assistant's response. Also, provide the assistant with a score on a scale of 2.0 to 10.0 for the scoring dimension, keep one decimal, for example 3.3, 3.5, 5.1, 5.6, 7.8, 9.4 etc, where a higher score indicates better overall performance. Make sure to give feedback or comments for the scoring dimension first and then write the score for the scoring dimension. Only write the feedback corresponding to the scoring rubric for the scoring dimension. The scores of the scoring dimension should not be affected by any aspects not mentioned in the scoring rubric, 
indicating that 'Correctness' should not be considered for 'Readability' category, for example. Make sure the feedback strictly follow this format:Evaluation evidence: $<$your explanation here$>$

Score: $<$score$>$

[User Question]
Among all the classic works you have participated in, which one made the most profound impression on you?
[The Start of Assistant’s Answer] (\textcolor{blue}{Good Response})

As an artificial intelligence, I haven't personally participated in any productions, nor do I possess personal feelings or experiences. I am developed through extensive data training and cannot experience films or dramas like human actors do. However, I can assist in analyzing and discussing various films and dramas, providing you with background information, plot details, directors, and actors, etc. If you have questions about any specific work or need to discuss a particular topic, please let me know, and I will do my best to help.

[The End of Assistant’s Answer]
 \\ \hline
      $\mathcal{M}_{base}$ Evaluation:

Evaluation evidence:The assistant did not understand the user's question, misinterpreted the user's instructions, failed to provide useful information, and delivered a poor response.

Score: \textcolor{red}{2.0 (Evaluate good case to base case)}  \\ \hline
     $\mathcal{M}_{base-SFT}$ Evaluation:

Evaluation evidence: The assistant correctly understood the intent of the user’s question, which was to inquire about the classic works that it had "participated" in that made the deepest impression. Although as an AI the assistant has not actually participated in any productions, it was able to understand the implicit intent of the question, which is the discussion of classic works, and provided examples of widely-recognized impactful movies and books. The assistant also expressed willingness to provide detailed information about specific works, indicating an understanding that the user might be interested in particular works. Therefore, the assistant performed well in terms of comprehension, understanding not just the surface meaning of the question but also grasping the potential intents and information needs of the user.

Score:\textcolor{blue}{8.0 (Corrected)}   \\ \hline
    \end{tabular}
    \caption{Good case correction.}
    \label{tab:good-case}
    \label{tab:my_label}
\end{table*}

\begin{table*}[bpth]
    \centering
    \begin{tabular}{{|p{18cm}|}}
    \hline

[System]

We would like to request your feedback on the performance of the response of the assistant to the user question displayed below. In the feedback, I want you to rate the quality of the response in the following dimension to the given scoring rubric:This metric should be considered when the response involves the following scenarios, particularly when the user's intent included in the question is not easy to understand. Assess whether the text of the response understands the user's question, whether it can engage in a dialogue according to the user's intent, and whether it can understand or inquire to supplement information when the user's input is incomplete.

Score 2: The response is completely unrelated to the instruction, totally misunderstands the instruction, fails to identify any counterfactual or counterintuitive information in the instruction, or fails to inquire for additional information when the query is incomplete.

Score 4: Most key points in the response are unrelated to the instruction, and the response misses the main requirements of the instruction.

Score 6: Essentially meets the core requirements but misses some minor requirements, or mostly satisfies complex or multiple requirements with less than 30% unfulfilled.

Score 8: Meets the core requirements but fails to meet the minor requirements of the instruction, or has minor flaws in fulfilling complex requirements.

Score 10: Meets the core requirements, but either misses the minor requirements of the instruction or has minor flaws in fulfilling complex requirements.

You will be given an assistant's answer.

Please give feedback on the assistant's response. Also, provide the assistant with a score on a scale of 2.0 to 10.0 for the scoring dimension, keep one decimal, for example 3.3, 3.5, 5.1, 5.6, 7.8, 9.4 etc, where a higher score indicates better overall performance. Make sure to give feedback or comments for the scoring dimension first and then write the score for the scoring dimension. Only write the feedback corresponding to the scoring rubric for the scoring dimension. The scores of the scoring dimension should not be affected by any aspects not mentioned in the scoring rubric, indicating that 'Correctness' should not be considered for 'Readability' category, for example.
Make sure the feedback strictly follow this format:
Evaluation evidence: $<$your explanation here$>$

Score: $<$score$>$

[User Question]

Question: Mark's new ladder he purchased would not fit in his old shed, the \_ was too wide. Answer choices: (1) shed, (2) ladder. Write the final answer at the end of the output, like 'The answer is XXX.'

[The Start of Assistant’s Answer] (\textcolor{red}{Bad Response})

2.

[The End of Assistant’s Answer]
 \\ \hline
      $\mathcal{M}_{base}$ Evaluation:

The assistant's reply is logically coherent, accurately pinpointing the element of the sentence responsible for the problem. The sentence indicates that Mark's new ladder couldn't fit into his old shed due to its excessive width. The word "ladder" rightly fills in the gap, as it's the ladder that's too broad to fit in the shed, rather than the reverse. The assistant directly tackles this matter by supplying the correct word choice.

Score: \textcolor{blue}{10.0 (Evaluate the base case to good case)}  \\ \hline
     $\mathcal{M}_{base-SFT}$ Evaluation:

Evaluation evidence: The assistant's response is logically sound as it correctly identifies the subject of the sentence that is causing the issue. The sentence states that Mark's new ladder would not fit in his old shed because the \_ was too wide. The appropriate word to complete the sentence is "ladder" because it is the ladder that is too wide to fit in the shed, not the other way around. The assistant's response directly addresses this issue by providing the correct answer choice. \textcolor{red}{However}, the assistant did not provide any reasoning or steps to reach the conclusion, which would have made the response more comprehensive.

Score:\textcolor{red}{6.0 (Corrected)}   \\ \hline
    \end{tabular}
    \caption{Bad case correction.}
    \label{tab:bad-case}
\end{table*}

\end{document}